\documentclass{article} 

\usepackage{microtype}
\usepackage{graphicx}
\usepackage{subfigure}
\usepackage{booktabs} 
\usepackage{gensymb}
\usepackage{svg}
\usepackage{amssymb}
\usepackage{multicol,multirow}
\usepackage{fixltx2e} 
\usepackage{tikz-cd}

\usepackage{hyperref}

\usepackage[accepted]{icml2021}
\usepackage{amsmath,amsthm}
\theoremstyle{definition}
\newtheorem{definition}{Definition}[section]
\theoremstyle{proposition}
\newtheorem{proposition}{Proposition}[section]
\usepackage{amsfonts}
\usepackage{mathrsfs}  

\usepackage{scrextend}

\usepackage[new]{old-arrows}
\newcommand{\sehookarrow}{\mathrel{\rotatebox[origin=c]{-45}{$\hookrightarrow$}}} 
\newcommand{\swhookarrow}{\mathrel{\rotatebox[origin=c]{-135}{$\hookrightarrow$}}} 

\icmltitlerunning{Z-GCNETs: Time Zigzags at Graph Convolutional Networks for Time Series Forecasting}

\begin{document}

\twocolumn[
\icmltitle{Z-GCNETs: Time Zigzags at Graph Convolutional Networks \\for Time Series Forecasting}



\icmlsetsymbol{equal}{}



\begin{icmlauthorlist}
\icmlauthor{Yuzhou Chen}{do,goo}
\icmlauthor{Ignacio Segovia-Dominguez}{ed,na}
\icmlauthor{Yulia R. Gel}{ed,goo}
\end{icmlauthorlist}

\icmlaffiliation{do}{Department of Statistical Science, Southern Methodist University, TX, USA}
\icmlaffiliation{ed}{Department of Mathematical Sciences, University of Texas at Dallas, TX, USA}
\icmlaffiliation{goo}{Energy Storage \& Distributed Resources Division, Lawrence Berkeley National Laboratory, CA, USA}
\icmlaffiliation{na}{NASA Jet Propulsion Laboratory,  CA, USA}

\icmlcorrespondingauthor{Yuzhou Chen}{yuzhouc@smu.edu}
\icmlcorrespondingauthor{Ignacio Segovia Dominguez}{ignacio.segoviadominguez@utdallas.edu}
\icmlcorrespondingauthor{Yulia R. Gel}{ygl@utdallas.edu}

\icmlkeywords{Machine Learning, ICML}

\vskip 0.3in
]



\printAffiliationsAndNotice{}  

\begin{abstract}
There recently has been a surge of interest in developing a new class of deep learning (DL) architectures that integrate an explicit time dimension as a fundamental building block of learning and representation mechanisms. In turn, many recent results show that topological descriptors of the observed data, encoding information on the shape of the dataset in a topological space at different scales, that is, persistent homology of the data, may contain important complementary information, improving both performance and robustness of DL. As convergence of these two emerging ideas,
we propose to enhance DL architectures with the most salient time-conditioned topological information of the data and introduce the concept of zigzag persistence into time-aware graph convolutional networks (GCNs). Zigzag persistence provides a systematic and mathematically rigorous framework to track the most important topological features of the observed data that tend to manifest themselves over time. To integrate the extracted time-conditioned topological descriptors into DL, we develop a new topological summary, zigzag persistence image, and derive its theoretical stability guarantees. 
We validate the new GCNs with a time-aware zigzag topological layer (Z-GCNETs), in application to traffic forecasting and Ethereum blockchain price prediction. Our results indicate that Z-GCNET outperforms 13 state-of-the-art methods on 4 time series datasets.

\end{abstract}

\section{Introduction}

Many real world phenomena are intrinsically dynamic by nature, and ideally neural networks, encoding the knowledge about the world should also be based on more explicit time-conditioned representation and learning mechanisms. However, most currently available deep learning (DL) architectures are inherently static and do not systematically integrate time-dimension into the learning process.
As a result, such model architectures often cannot reliably, accurately and on time learn
many salient time-conditioned characteristics of complex interdependent systems, often resulting in outdated decisions and requiring frequent model updates.

In turn, in the last few years we observe an increasing interest to integrate deep neural network architectures with persistent homology representations of the learned objects, typically in a form of some topological layer in DL~\cite{hofer2019learning,carriere2020perslay, carlsson2020topological}. Such persistent homology representations allow us to extract and learn descriptors of the object {\it shape}. (By shape here we broadly understand data characteristics that are invariant under continuous transformations such as bending, stretching, and compressing.) 
Such interest in combining persistent homology representations with DL is explained by the complementary multi-scale information topological descriptors deliver about the underlying objects, and
higher robustness of these salient object characterisations to perturbations. 

Here we take the first step toward merging the two directions. To enhance DL with the most salient {\it time-conditioned topological} information, we introduce the concept of zigzag persistence into time-aware DL. Building on the fundamental results on quiver representations, zigzag persistence 
studies properties of topological spaces
which are connected via inclusions going in both directions~\cite{carlsson2010zigzag, tausz2011applications,carlsson2019persistent}. Such generalization of ordinary persistent homology allows us to track topological properties of time-conditioned objects by extracting and tracking  salient {\it time-aware topological features} through time-ordered inclusions. We propose to summarize the extracted time-aware zigzag persistence in a form of zigzag persistence images and then integrate the resulting information as a learnable time-aware zigzag layer into GCN.  

The key novelty of our paper can be summarized as follows:
\begin{itemize}
    \item This is the first approach bridging
    time-conditioned DL with time-aware persistent homology representations of the data.
    
    \item We propose a new vectorized summary for time-aware persistence, namely, {\it zigzag persistence image} and discuss its theoretical stability guarantees. 
    
    \item We introduce the concepts of time-aware zigzag persistence into learning time-conditioned graph structures and develop a 
    zigzag topological layer (Z-GCNET) for time-aware graph convolutional networks (GCNs).
    
    \item Our experiments on application Z-GCNET to traffic forecasting and Ethereum blockchain price prediction show that 
Z-GCNET surpasses 13 state-of-the-art methods on 4 benchmark datasets, both in terms of accuracy and robustness.
    \end{itemize}

\section{Related Work}

{\bf Zigzag Persistence} is yet an emerging tool in applied topological data analysis, but many recent studies have already shown its high utility in such diverse applications as brain sciences~\cite{chowdhury2018importance}, imagery classification~\cite{adams2020torus}, cyber-security of mobile sensor networks~\cite{adams2015evasion, gamble2015coordinate}, and characterization of flocking and swarming behavior in biological sciences~\cite{corcoran2017modelling, kim2020analysis}.
An alternative to zigzag but a closely related approach to assess properties of time-varying data with persistent homology, namely, crocker stacks, has been recently suggested by~\citet{xian2020capturing}, though the crocker stacks representations are not learnable in DL models.    
While zigzag has been studied in conjunction with dynamic systems~\cite{tymochko2020hopf} and time-evolving point clouds~\cite{corcoran2017modelling}, till now, the utility of zigzag persistence remains untapped not only in conjunction with GCNs but with any other DL tools.    

{\bf Time series forecasting}
From a deep learning perspective, Recurrent Neural Networks (RNNs) are natural methods to model time-dependent datasets~\cite{RRN:Yong:2019}. In particular, the stable architecture of Long Short Term Memories (LSTMs), and its variant called Gate Recurrent Unit (GRU), solves the gradient instability of predecessors and adds extra flexibility due to their memory storage and forget gates. The ability of LSTM and GRU to selectively learn historical patterns awaken an interest among researchers and major companies to solve a variety of time-dependant machine learning problems~\cite{LSTMdisease:Sangwon:2018,LSTMcompanies,GRU:Yuan:2019,GRU:Shin:2020}. Although most of variants of LSTM architecture performs similarly well in large scale studies, see \cite{LSTM:Greff:2017}, GRU models has fewer parameters and, in general, perform similarly well as LSTM \cite{GRU-LSTM:Gao:2020}. Applications of RNN are limited by the underlying structure of the input data, these methods are not designed to handle data from non-Euclidean spaces, such as graphs and manifolds.

{\bf Graph convolutional networks} To overcome the limitations of traditional convolution on graph structured data, graph convolution-based methods~\cite{DefferrardNIPS,kipf2016semi,velivckovic2017graph} are proposed to explore both global and local structures. GCNs usually consists of graph convolution layers which extract the edge characteristics between neighbor nodes and aggregate feature information from neighborhood via graph filters. In addition to convolution, there has been a surge of interest in applying GCNs time series forecasting tasks~\cite{yuijcai2018, yao2018deep, yan2018spatial,guo2019attention, weber2019anti,pareja2020evolvegcn}. Although these methods have achieved state-of-the-art performance in traffic flow forecasting, human action recognition, and anti-money laundering regulation, the design of spatial temporal graph convolution network framework is mostly based on modeling spatial-temporal correlation in terms of feature-level and pre-defined graph structure. 
%
\section{Time-Aware Topological Signatures of Graphs}

{\bf Spatio-temporal Data as Graph Structures} The spatial-temporal networks can be represented as a sequence of discrete snapshots, $\{\mathcal{G}_1, \mathcal{G}_2, \dots, \mathcal{G}_{T}\}$, where $\mathcal{G}_t = \{\mathcal{V}, \mathcal{E}_t, W_t$\} represents the graph structure at time step $t$, $t=1,\ldots, T$. In spatial network $\mathcal{G}_t$, $\mathcal{V}$ is a node set with cardinality $|\mathcal{V}|$ of $N$ and $\mathcal{E}_t \subseteq \mathcal{V} \times \mathcal{V}$ is an edge set. A nonnegative symmetric $N\times N$-matrix $W_t$ with entries $\{\omega^t_{ij}\}_{1\leq i,j\leq N}$ represents the adjacency matrix of $\mathcal{G}_t$, that is, $\omega^t_{ij} > 0$ for any $e^t_{i j} \in \mathcal{E}_t$ and $\omega^t_{ij} = 0$, otherwise. Let $F, F\in \mathbb{Z}_{> 0}$ be the number of different node features associated each node $v\in \mathcal{V}$. Then, a $N \times F$ feature matrix $\boldsymbol{X}_t$ serves as an input to the framework of time series process modeling.

{\bf Background on Persistent Homology} Persistent homology is a mathematical machinery to extract the intrinsic shape properties of $\mathcal{G}$ that are invariant under continuous transformations such as bending, stretching, and twisting. The key idea is, based on some appropriate scale parameter, to associate $\mathcal{G}$ with a graph filtration $\mathcal{G}^{1} \subseteq \ldots \subseteq \mathcal{G}^{n}=\mathcal{G}$ and then to equip each $\mathcal{G}^i$ with an abstract simplicial complex $\mathscr{C}(\mathcal{G}^{i})$, $1\leq i\leq n$, yielding a filtration of complexes $\mathscr{C}(\mathcal{G}^{1}) \subseteq \ldots \subseteq \mathscr{C}(\mathcal{G}^{n})$. Now, we can systematically and efficiently  track evolution of various patterns such as connected components, cycles, and voids throughout this hierarchical sequence of complexes. Each topological feature, or $p$-hole (e.g., number of connected components and voids), $0\leq p\leq d$, 
is represented by a unique pair $(i_{b}, j_{d})$, where
birth $i_{b}$ and death $j_{d}$ are the scale parameters at which the feature first appears and disappears, respectively. 
The lifespan of the feature is defined as
$i_{d}-j_{b}$.  The extracted topological information can be then summarized as a persistence diagram ${\rm{Dgm}}=\{(i_{b}, j_{d}) \in \mathbb{R}^{2} | i_{b}<j_{d}\}$. Multiplicity of a point $(i_{b}, j_{d}) \in \mathcal{D}$ is the number of $p$-dimensional topological features ($p$-holes) that are born and die at $i_{b}$ and $j_{b}$, respectively. Points at the diagonal $\rm{Dgm}$ are taken with infinite multiplicity.
The idea is then to evaluate topological features that
persist (i.e. have longer lifespan) over the complex filtration and, hence, are likelier to contain important structural information on the graph.

Finally, a filtration of the weighted graph $\mathcal{G}$ can be constructed in multiple ways. For instance, (i) we can select a scale parameter as edge weight and, as an abstract simplicial complex
$\mathscr{C}$ on $\mathcal{G}$, consider a Vietoris–Rips (VR) complex
$\mathscr{VR}^{\nu_{*}}(\mathcal{G})= \{\mathcal{G}^{'} \subseteq \mathcal{G}| diam(\mathcal{G}^{'})\leq \nu_{*}\}$, that is, 
$\mathscr{VR}^{\nu_{*}}(\mathcal{G})$ consists of nodes with a shortest weighted path of at most $\nu_{*}$. Hence, for a set of scale thresholds
$\nu_{1}\leq \ldots \nu_{n}$, we obtain a VR filtration $\mathscr{VR}^{{1}} \subseteq \ldots \subseteq \mathscr{VR}^{{n}}$.
Alternatively, (ii) we can consider a sublevel filtration induced by a continuous function $f$ defined on nodes of $\mathcal{G}$. Let
$f:V \rightarrow \mathbb{R}$ and  $\nu_1<\nu_2<\ldots <\nu_n$ be a sequence of sorted filtered values, then $\mathscr{C}^{i}=\{\sigma \in \mathscr{C}: \max_{v\in \sigma}f(v)\leq \nu_i\}$.
Note that a VR filtration (i) is a subcase of sublevel filtration (ii) with $f$ being the diameter function~\cite{adams2017persistence, bauer2019ripser}.

{\bf Time-Aware Zigzag Persistence} Since our primary aim is to
assess interconnected evolution of multiple time-conditioned objects, the developed methodology for tracking topological and geometric properties of these objects shall ideally account for their intrinsically dynamic nature.  
We address this goal by introducing the concept of {\it zigzag} persistence into GNN. Zigzag persistence is a generalization of persistent homology proposed by~\cite{carlsson2010zigzag} and 
provides a systematic and mathematically rigorous framework to track the most important topological features of the data persisting over time. 

Let $\{\mathcal{G}_t\}_{1}^T$ be a sequence of networks observed over time. The key idea of zigzag persistence is to evaluate  pairwise compatible topological features in this time-ordered sequence of networks. First, we define
a set of network inclusions over time
\begin{align*}
\label{net_incl}
\mathcal{G}_1 & & & & \mathcal{G}_{2}& & & & \mathcal{G}_{3} & \;\;\ldots \\
  & \sehookarrow & & \swhookarrow &  & \sehookarrow & & \swhookarrow & & \;\;\;\;\;\;\; ,\\
   & & \mathcal{G}_1 \cup \mathcal{G}_{2}& &  & & \mathcal{G}_2 \cup \mathcal{G}_{3} & & &
\end{align*}
where $\mathcal{G}_k \cup \mathcal{G}_{k+1}$ is defined as a graph with a node set $V_k \cup V_{k+1}$ and an edge set $E_k \cup E_{k+1}$. 
Second, we fix a scale parameter $\nu_{*}$ and build a zigzag diagram
of simplicial complexes for the given $\nu_{*}$
over the constructed set of network inclusions
\begin{align*}
\mathscr{C}(\mathcal{G}_1, \nu_{*}) & & & & \mathscr{C}(\mathcal{G}_{2}, \nu_{*})& & & 
\\
  & \sehookarrow & & \swhookarrow &  & \sehookarrow & 
  \\
   & & \mathscr{C}(\mathcal{G}_1 \cup \mathcal{G}_{2}, \nu_{*})& &  & & \ldots 
\end{align*}
Using the zigzag filtration for the given $\nu_{*}$, we can track birth and death of each topological feature over $\{\mathcal{G}_t\}_{1}^T$ as time points $t_b$ and $t_d$, $1\leq t_b\leq t_d\leq T$, respectively. Similarly to a non-dynamic case, we can extend the notion of persistence diagram for the analysis of topological characteristics of time-varying data delivered by the zigzag persistence.

\begin{definition}[Zigzag Persistence Diagram (ZPD)]
\label{zigzagPD}
Let $t_b$ and $t_d$ be time points, when a topological feature first appears (i.e., is born) and disappears (i.e., dies) in the time period $[1,T]$ over the zigzag diagram of simplicial complexes for a fixed scale parameter $\nu_{*}$, respectively.
If the topological feature first appears in $\mathscr{C}(\mathcal{G}_k, \nu_{*})$, $t_b=k$; if it first appears in $\mathscr{C}(\mathcal{G}_k \cup \mathcal{G}_{k+1}, \nu_{*})$, $t_b=k+1/2$. Similarly, if a topological feature last appears in $\mathscr{C}(\mathcal{G}_k, \nu_{*})$, $t_d=k$; and if it last appears in $\mathscr{C}(\mathcal{G}_k \cup \mathcal{G}_{k+1}, \nu_{*})$, $t_d=k+1/2$.
A multi-set of points in $\mathbb{R}^2$ $\rm{DgmZZ}_{\nu_{*}}=\{(t_{b}, t_{d}) \in \mathbb{R}^{2} | t_{b}<t_{b}\}$ for a fixed
$\nu_{*}$ is called a {\it zigzag persistence diagram (ZPD)}.
\end{definition}
 Inspired by the notion of a persistent image as a summary of ordinary persistence~\cite{adams2017persistence}, to input topological information summarized by ZPD into a GNN, we propose a  representation of ZPD as {\it zigzag persistence image} (ZPI). 
 ZPI is a finite-dimensional vector representation of a ZPD and can be computed through the following steps:
\begin{itemize}
\item\textit{Step 1:} Map a zigzag persistence diagram $\rm{DgmZZ}_{\nu_{*}}$ to an integrable function $\rho_{\rm{DgmZZ}_{\nu_{*}}}: \mathbb{R}^2 \rightarrow \mathbb{R}^{2}$, called a {\it zigzag persistence surface}. The zigzag persistence surface is given by sums of weighted Gaussian functions that are centered at each point in $\rm{DgmZZ}_{\nu_{*}}$.
\item \textit{Step 2:} Perform a discretization and linearization of a subdomain of zigzag persistence surface $\rho_{\rm{DgmZZ}_{\nu_{*}}}$ in a grid.
\item \textit{Step 3:} The ZPI, i.e., a matrix of pixel values, can be obtained by subsequent integration over each grid box.
\end{itemize}
\begin{figure}[b!]
\centering
\includegraphics[width=0.45\textwidth]{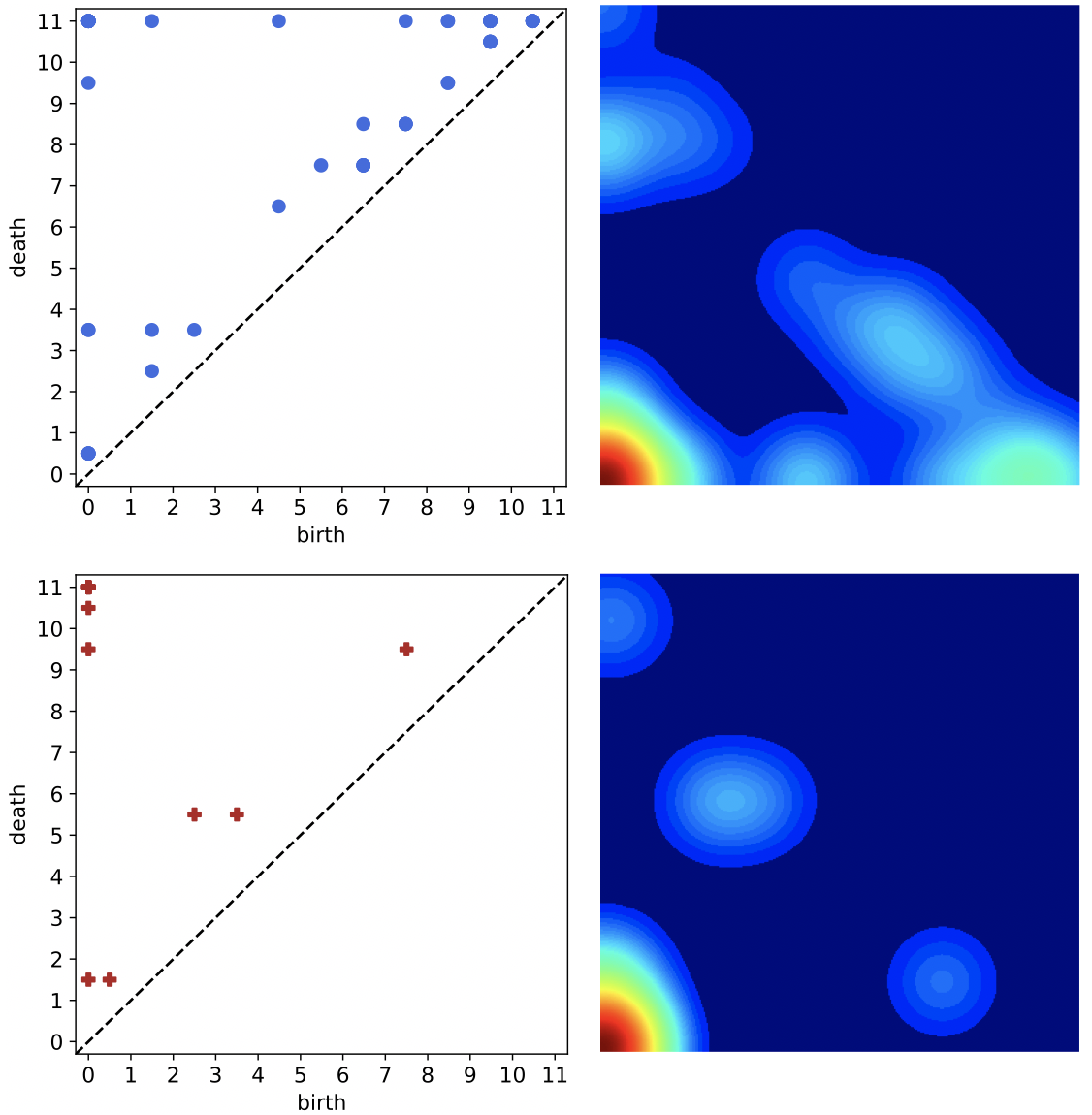}
\caption{Illustrations of 0- and 1-dimensional ZPD and 0- and 1-dimensional ZPI for PeMSD4 dataset using the sliding window size $\tau = 12$, i.e., dynamic network with 12 graphs. Upper part shows the 0-dimensional ZPD and ZPI whilst the lower part is the 1-dimensional ZPD and ZPI.\label{ZPD_ZPI}}
\end{figure}
The value of each pixel $z \in \mathbb{R}^{2}$ within a ZPI is defined as:
\begin{align*}
\rm{ZPI}_{\nu_{*}}(z) =\iint\limits_{z}\sum_{\mu \in \rm{DgmZZ}^{\prime}_{\nu_{*}}} g\left(\mu\right)\exp{\bigg\{-\frac{||z-\mu||^2}{2\vartheta^2}\bigg\}} dz_x d z_y,
\end{align*}
where $\rm{DgmZZ}^{\prime}_{\nu_{*}}$ is the transformed multi-set in $\rm{DgmZZ}_{\nu_{*}}$, i.e., $\rm{DgmZZ}^{\prime}_{\nu_{*}} (x,y) = (x,y-x)$; $g(\mu)$ is a weighting function with mean $\mu = (\mu_x, \mu_y) \in \mathbb{R}^2$ and variance $\vartheta^2$, which depends on the distance from the diagonal. Here the zigzag persistence surface is defined by $\rho_{\rm{DgmZZ}_{\nu_{*}}} = \sum_{\mu \in ZPD^{\prime}}g\left(\mu\right)\exp{\bigl\{-{||z-\mu||^2}/{2\vartheta^2}\bigr\}}$.

\begin{proposition}
Let $g:\mathbb{R}^2\to \mathbb{R}$ be a non-negative continuous and piece-wise differentiable function. 
Let $\rm{DgmZZ}_{\nu_{*}}$ be a zigzag persistence diagram
for some fixed scale parameter
$\nu_{*}$, and let $\rm{ZPI}_{\nu_{*}}$ be its corresponding zigzag persistence image.
Then, $\rm{ZPI}_{\nu_{*}}$ is stable with respect to the Wasserstein-1 distance between zigzag persistence diagrams.
\end{proposition}

Tracking evolution of topological patterns in these sequences of time-evolving graphs allows us to glean insights into which properties of the observed time-conditioned objects, e.g., traffic data or Ethereum transaction graphs, tend to persist over time and, hence, are likelier to play a more important role in predictive tasks. 


\section{Z-GCNETs}
Given the graph $\mathcal{G}$ and graph signals $\boldsymbol{X}^{\tau} = \{\boldsymbol{X}_{t-\tau}, \dots, \boldsymbol{X}_{t-1}\} \in \mathbb{R}^{\tau \times N \times F}$ of $\tau$ past time periods (i.e. window size $\tau$; where $\boldsymbol{X}_i \in \mathbb{R}^{N \times F}$ and $i \in \{t-\tau, \dots, t-1\}$), we employ a model targeted on multi-step time series forecasting. 
That is, given the windows size $\tau$ of past graph signals and the ahead horizon size $h$,
our goal is to learn a mapping function which maps the historical data $\{\boldsymbol{X}_{t-\tau}, \dots, \boldsymbol{X}_{t-1}\}$ into the future data $\{\boldsymbol{X}_{t}, \dots, \boldsymbol{X}_{t+h}\}$.
\begin{figure*}[t!]
    \centering
    \includegraphics[width=0.95\textwidth]{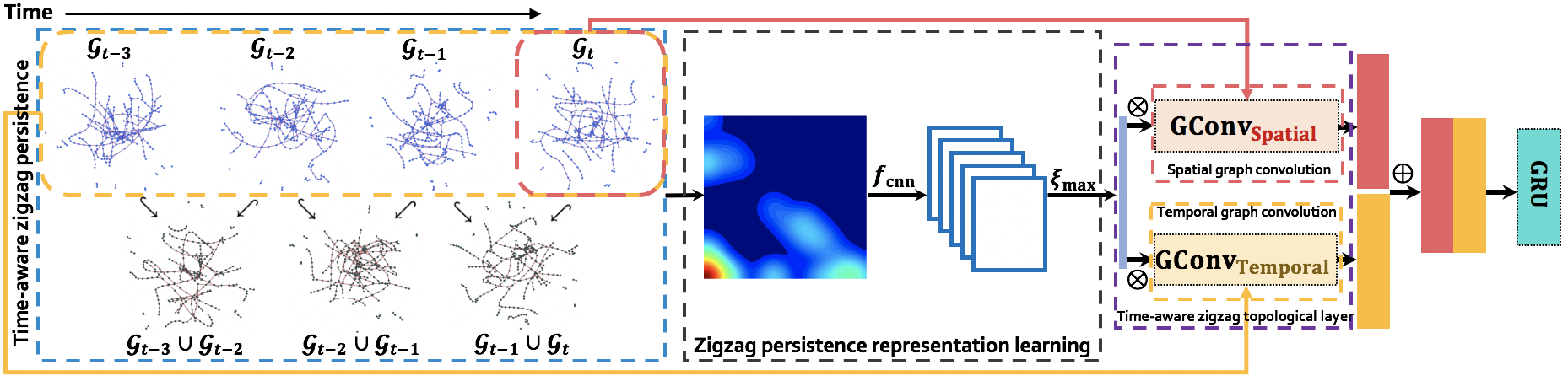}
    \caption{The architecture of Z-GCNETs. Given a sliding window, e.g. $(\mathcal{G}_{t-3}, \dots, \mathcal{G}_t)$, we extract zigzag persistence image (ZPI) based on zigzag filtration. For the ZPI $\in \mathbb{R}^2$ with the shape of $p \times p$, Z-GCNETs first learn the topological features of the ZPI through CNN-based framework, and then apply global max-pooling to obtain the maximum values among pooled activation maps. The output of zigzag persistence representation learning is decoded into spatial graph convolution and temporal graph convolution, where the inputs of spatial graph convolution and temporal graph convolution are current timestamps, e.g. $\mathcal{G}_t$ in red dashed box, and the sliding window (i.e., $(\mathcal{G}_{t-3}, \dots, \mathcal{G}_t)$ in yellow dashed box) respectively. After graph convolution operations, features from time-aware zigzag topological layer are combined and moved to GRU to perform forecasting. Symbol $\otimes$ represents dot product whilst $\oplus$ denotes combination.\label{flowchart}}
\end{figure*}

\textbf{Laplacianlink} In spatial-temporal domain, the topology of graph may have different structure at different points in time. In this paper, we use the self-adaptive adjacency matrix~\cite{graphwavenet} as the normalized Laplacian by trainable node embedding dictionaries $\boldsymbol{\phi} \in \mathbb{R}^{N \times c}$, i.e., $\boldsymbol{L} = softmax(ReLU(\boldsymbol{\phi}\boldsymbol{\phi}^{\top}))$, where the dimension of embedding $c \geq 1$. Although introducing node embedding dictionaries allows capture hidden spatial dependence information, it cannot sufficiently capture the global graph information and the similarity between nodes. To overcome the limits and explore neighborhoods of nodes at different depths, we define a novel polynomial representation for Laplacian based on positive powers of the Laplacian matrix. In this work, Laplacianlink $\Tilde{\boldsymbol{L}}$ is formulated as:
\begin{align}
    \Tilde{\boldsymbol{L}} = \left[\boldsymbol{I}, \boldsymbol{L}, \boldsymbol{L}^2, \dots, \boldsymbol{L}^{K}\right] \in \mathbb{R}^{N \times N \times (K+1)},
\end{align}
where $K \geq 1$, $\boldsymbol{I} \in \mathbb{R}^{N \times N}$ represents the identity matrix, and $\boldsymbol{L}^{k} \in \mathbb{R}^{N \times N}$, with $0\leq k \leq K$, denotes the power series of normalized Laplacian.

By {\it linking} (i.e., stacking) the power series of normalized Laplacian, we build a diffusion formalism to accumulate neighbors' information of different power levels. Hence, each node will successfully exploit and propagate spatial-temporal correlations after spatial and temporal graph convolutional operations. 

\textbf{Spatial graph convolution}
To model the spatial network $\mathcal{G}_t$ at timestamp $t$ with its node feature matrix $\boldsymbol{X}_t$, we define the spatial graph convolution as multiplying the input of each layer with the Laplacianlink $\Tilde{\boldsymbol{L}}$, which is then fed into the trainable projection matrix $\boldsymbol{\Theta} = \boldsymbol{\phi}\boldsymbol{V}$ (where $\boldsymbol{V}$ stands for the trainable weight). In spatial-temporal graph modeling, we prefer to use weight sharing in matrix factorization rather than directly assigning a trainable weight matrix in order not only to avoid the risk of over-fitting but also to reduce the computational complexity. We compute the transformation in spatial domain, in each layer, as follows:
\begin{align}
    \boldsymbol{H}^{(\ell)}_{i,\mathcal{S}} = (\Tilde{\boldsymbol{L}}\boldsymbol{H}^{(\ell-1)}_{i,\mathcal{S}})^{\top}\boldsymbol{\phi}\boldsymbol{V},
\end{align}
where $\boldsymbol{\phi} \in \mathbb{R}^{N \times c}$ is the node embedding and $\boldsymbol{V} \in \mathbb{R}^{c \times (K+1) \times C_{in} \times ({C_{out}}/{2})}$ is the trainable weight ($C_{in}$ and $C_{out}$ are the number of channels in input and output, respectively). $\boldsymbol{H}^{(\ell-1)}_{i,\mathcal{S}} \in \mathbb{R}^{N \times ({C_{in}}/{2})}$ is the matrix of activations of spatial graph convolution to the $\ell$-th layer and $\boldsymbol{H}^{(0)}_{i,\mathcal{S}} = \boldsymbol{X}_i$. As a result, all information regarding the $\ell$-layered input at time $i$ are reflected in the latest state variable. 

\textbf{Temporal graph convolution}
In addition to spatial domain, the nature of spatial-temporal networks includes temporal relationships among consistent spatial networks. To catch the temporal correlation patterns of nodes, we choose longer window size (i.e., by using the entire sliding window as input) and apply temporal graph convolution to graph signals in sliding window $\boldsymbol{X}^{\tau}$. The mechanism has several excellent properties: (i) there is no need to select a particular size of nested sliding window, i.e., does not introduce  additional computational complexity, (ii) temporal correlation patterns can be well captured and evaluated by (long) window sizes, whereas short window sizes (i.e., nested sliding window) are likely to be bias and noise, and (iii) the sliding window $\boldsymbol{X}^{\tau}$ maximizes the efficiency of estimating temporal correlations. The temporal graph convolution is presented in the form:
\begin{align}
\boldsymbol{H}^{(\ell)}_{i,\mathcal{T}}  = \left((\Tilde{\boldsymbol{L}}\boldsymbol{H}^{(\ell-1)}_{i,\mathcal{T}})^{\top}\boldsymbol{\phi}\boldsymbol{U}^{(\ell-1)}\right)\boldsymbol{Q},
\end{align}
where $\boldsymbol{U}^{(\ell)}\in \mathbb{R}^{c \times C_{in} \times ({C_{out}}/{2})}$ is trainable weight and $\boldsymbol{U}^{(0)} \in \mathbb{R}^{c \times F \times ({C_{out}}/{2})}$, $\boldsymbol{Q} \in \mathbb{R}^{\tau \times 1}$ is the trainable projection vector in temporal graph convolutional layer, and $\boldsymbol{H}^{(\ell-1)}_{i,\mathcal{T}} \in \mathbb{R}^{\tau \times N \times ({C_{in}}/{2})}$ is the hidden matrix fed to the $\ell$-th layer and $\boldsymbol{H}^{(0)}_{i,\mathcal{T}} = \boldsymbol{X}^{\tau} \in \mathbb{R}^{\tau \times N \times F}$.

\textbf{Time-aware zigzag topological layer}
To learn the topological features across a range of spatial and temporal scales, we extend the CNN model to be used along with ZPI. In this research, we present a framework to aggregate the topological persistent features into the feature representation learned from GCN. Let $\text{ZPI}^{\tau}$ denotes the ZPI based on the sliding window $\boldsymbol{X}^{\tau}$. 
(Here for brevity we suppress dependence of ZPI on a scale parameter
$\nu_{*}$.)
We design the time-aware zigzag topological layer to (i) extract and learn the spatial-temporal topological features contained in ZPI, (ii) aggregate transformed information from (spatial or temporal) graph convolution and spatial-temporal topological information from zigzag persistence module, and (iii) mix spatial-temporal and spatial-temporal topological information.
The information's extraction, aggregation, and combination processes are expressed as: 
\begin{align}
\begin{split}
\boldsymbol{Z}^{(\ell)} &= \xi_{\text{max}}\left(f^{(\ell)}_{\text{cnn}}(\text{ZPI}^{\tau})\right),\\
\boldsymbol{S}_{i}^{(\ell)} & = \boldsymbol{H}^{(\ell)}_{i,\mathcal{S}}\boldsymbol{Z}^{(\ell)},\\
\boldsymbol{T}_{i}^{(\ell)}& = \boldsymbol{H}^{(\ell)}_{i,\mathcal{T}}\boldsymbol{Z}^{(\ell)},\\
\boldsymbol{H}^{(\ell)}_{i,out} &= \text{COMBINE}^{(\ell)}(\boldsymbol{S}_{i}^{(\ell)}, \boldsymbol{T}_{i}^{(\ell)}),\\
\end{split}
\end{align}
where $f_{\text{cnn}}^{(\ell)}$ represents the convolutional neural network (CNN) in the $\ell$-th layer, $\xi_{\text{max}}(\cdot)$ denotes global max-pooling operation, $\boldsymbol{Z}^{(\ell)} \in \mathbb{R}^{({C_{out}}/{2})}$ is the learned zigzag persistence representation from CNN, $\boldsymbol{S}^{(\ell)}_i \in \mathbb{R}^{N \times ({C_{out}}/{2})}$ is the aggregated $\text{spatial}^2$-temporal representation, $\boldsymbol{T}^{(\ell)}_i \in \mathbb{R}^{N \times ({C_{out}}/{2})}$ is the aggregated spatial-$\text{temporal}^2$ representation, and the output of time-aware zigzag topological layer $\boldsymbol{H}^{(\ell)}_{i,out} \in \mathbb{R}^{N \times C_{out}}$ combines hidden states $\boldsymbol{S}^{(\ell)}_i$ and $\boldsymbol{T}^{(\ell)}_i$ at time $i$. 

\textbf{GRU with time-aware zigzag topological layer} GRU is a variant of LSTM network. Compared with LSTM, GRU has a simpler structure, fewer training parameters, and more easily overcome vanishing and exploding gradient problems. The feed forward propagation of GRU with time-aware zigzag topological layer is recursively conducted as:
\begin{align}
\begin{split}
\boldsymbol{z}_i &= \varphi\left(\boldsymbol{W}_z\left[\boldsymbol{O}_{i-1}, \boldsymbol{H}_{i,out}\right] + \boldsymbol{b}_z\right),\\
\boldsymbol{r}_i & = \varphi\left(\boldsymbol{W}_r\left[\boldsymbol{O}_{i-1}, \boldsymbol{H}_{i,out}\right]+ \boldsymbol{b}_r\right),\\
\Tilde{\boldsymbol{O}}_i & = \tanh{\left(\boldsymbol{W}_o\left[\boldsymbol{r}_i \odot \boldsymbol{O}_{i-1}, \boldsymbol{H}_{i,out}\right]+ \boldsymbol{b}_o\right)},\\
\Tilde{\boldsymbol{O}}_i & = \boldsymbol{z}_t \odot \boldsymbol{O}_{i-1} + (1- \boldsymbol{z}_t) \odot \Tilde{\boldsymbol{O}}_i,\\
\end{split}
\end{align}
where $\varphi(\cdot)$ is a non-linear function, i.e., the ReLU function; $\odot$ is the elementwise product; $\boldsymbol{z}_i$ and $\boldsymbol{r}_i$ are update gate and reset gate, respectively; $\boldsymbol{b}_z$, $\boldsymbol{b}_r$, $\boldsymbol{b}_o$, $\boldsymbol{W}_z$, $\boldsymbol{W}_r$, and $\boldsymbol{W}_o$ are trainable parameters; $\left[\boldsymbol{O}_{i-1}, \boldsymbol{H}_{i,out}\right]$ and $\boldsymbol{O}_i$ are the input and output of GRU model, respectively. In this way, Z-GCNETs contains structural, temporal, and topological information.

\section{Experiments}

\subsection{Datasets}
We consider two types of networks (i) traffic network and (ii) Ethereum token network. Statistical overview of all datasets is given in Table~\ref{dataset}. We now describe the detailed construction of traffic and Ethereum transaction networks as follows
\begin{table}[t]
\centering
\setlength\tabcolsep{1.2pt}
\caption{Summary of datasets used in time series forecasting tasks. $[\dagger]$ means the average number of edges in transportation networks under threshold $\nu_{*}$.\label{dataset}}
\begin{tabular}{lccc}
\toprule
\textbf{Dataset} & \textbf{\# Nodes} & \textbf{Avg \# edges} & \textbf{Time range}\\
\midrule
Bytom &100&9.98&{\small 27/07/2017 - 07/05/2018}\\
Decentraland &100&16.94&{\small 14/10/2017 - 07/05/2018}\\
PeMSD4 &307&316.10$^\dagger$&{\small 01/01/2018 - 28/02/2018}\\
PeMSD8 &170&193.53$^\dagger$&{\small 01/07/2016 - 31/08/2016}\\
\bottomrule
\end{tabular}
\end{table}
(i) The freeway Performance Measurement System (PeMS) data sources (i.e., PeMSD4 and PeMSD8)~\cite{chen2001freeway} collects real time traffic data in California. Both PeMSD4 and PeMSD8 datasets are aggregated to 5 minutes, therefore there are overall 16,992 and 17,856 data points in PeMSD4 and PeMSD8, respectively. In the traffic network, the node is represented by the loop detector which can detect real time measurement of traffic conditions and the edge is a freeway segment between two nearest nodes. Thus, the node feature matrix of traffic network $X_t \in \mathbb{R}^{N \times 3}$ denotes that each node has 3 features (i.e., flow rate, speed, and occupancy) at time $t$. To capture both spatial and temporal dependencies, we reconstruct the traffic graph structure $\mathcal{G}_t = \{\mathcal{V}, \mathcal{E}, W^{\nu_{*}}_t\}$ at time $t$. Here, we define the {\it right censoring} weight $W^{\nu_{*}}_t$
\begin{align}
    \omega^{\nu_{*}}_{t, uv} =
    \begin{cases}
    w_{t, uv} & (u,v) \in \mathcal{E} \text{ and } w_{t, uv} \leq \nu_{*}\\
    0 & (u,v) \in \mathcal{E} \text{ and } w_{t, uv} > \nu_{*}\\
    0 & (u,v) \notin \mathcal{E}
    \end{cases},
\end{align}
where $w_{t, uv} = e^{{-||x_{t, u} - x_{t, v}||^2}/{\gamma}}$ is based on the Radial Basis Function (RBF). To investigate the topology of weighted graph, the traffic graph structure $\mathcal{G}_t$ is obtained via sub-level sets of the weight function, that is, we restrict to the final graph keep all edges of weights $\omega^{\nu_{*}}_{t, uv}$ below or equal to threshold $\nu_{*}$ and therefore threshold $\nu_{*}$ makes a difference in the topology of resulting graphs at different observation points. In experiments, we assign parameter $\gamma = 1.0$ to RBF and set the thresholds in PeMSD4 and PeMSD8 to $\nu_{*} = 0.5$ and $\nu_{*} = 0.3$, respectively.
\begin{table*}
\caption{Forecasting performance comparison of different approaches on PeMSD4 and PeMSD8 datasets.\label{forecast_result}}
\centering
\begin{tabular}{lcccccc}
\toprule
\multirow{2}{*}{\textbf{Model}}& \multicolumn{3}{c}{\textbf{PeMSD4}} & \multicolumn{3}{c}{\textbf{PeMSD8}}
\\
\cmidrule(lr){2-4}\cmidrule(lr){5-7}
           & MAE & RMSE & MAPE & MAE & RMSE & MAPE \\
\midrule
HA& 38.03 & 59.24 & 27.88\%& 34.86 & 52.04 & 24.07\% \\
VAR~\cite{hamilton2020time}& 24.54 & 38.61 &17.24\% & 19.19 & 29.81 & 13.10\%\\
FC-LSTM~\cite{sutskever2014sequence}& 26.77  & 40.65 & 18.23\% &23.09  &35.17  &14.99\%\\
GRU-ED~\cite{cho2014learning}& 23.68 & 39.27 & 16.44\%& 22.00 & 36.23 & 13.33\% \\
DSANet~\cite{huang2019dsanet}& 22.79  & 35.77 & 16.03\% & 17.14 & 26.96 & 11.32\%\\
DCRNN~\cite{li2017diffusion}& 21.22 & 37.23 & 14.17\%& 16.82 & 26.36 &10.92\%\\
STGCN~\cite{stgcn}&  21.16& 35.69 & 13.83\%& 17.50 & 27.09 &11.29\%\\
GraphWaveNet~\cite{graphwavenet} & 28.15 & 39.88 & 18.52\% & 20.30 & 30.82 & 13.84\% \\
ASTGCN~\cite{guo2019attention}& 22.93 & 34.33 & 16.56\%& 18.25 & 28.06 &11.64\%\\
MSTGCN~\cite{guo2019attention}&23.96& 37.21&14.33\% &19.00& 29.15& 12.38\% \\
STSGCN~\cite{song2020spatial}& 21.19 & 33.69 &13.90\% & 17.13 & 26.86 & 10.96\%\\
AGCRN~\cite{bai2020adaptive}&19.83  &32.30  &12.97\% & 15.95 & 25.22 & 10.09\%\\
LSGCN~\cite{LSGCN} &21.53& 33.86& 13.18\%& 17.73&26.76&11.20\%\\
\midrule
\textbf{Z-GCNETs (ours)} & {\bf 19.50} & {\bf 31.61} &{\bf 12.78\%} & {\bf 15.76} & {\bf 25.11}  & {\bf 10.01\%} \\
\bottomrule
\end{tabular}
\end{table*}
(ii) The Ethereum blockchain was developed in 2014 to implement Smart Contracts, which are used to create and sell digital assets on the network\footnote{\url{Ethereum.org}}. In particular, token assets are specially valuable because each token naturally represents a network layer with the same nodes, i.e., addresses of users, appearing in the networks, i.e., layers, of multiple tokens \cite{Tokens:Angelo:2020}. For our experiments, we extract two token networks with more than \$100M in market value\footnote{\label{priceEther}\url{EtherScan.io}}, Bytom and Decentraland tokens, from the publicly available Ethereum blockchain. We focus our analysis on the dynamic network generated by the daily transactions on each token network, and historical daily closed prices\footref{priceEther}. Since each token has different creation date\footnote{End date: May 7, 2018 }, Bytom dynamic network contains 285 nets whilst Decentraland dynamic network has 206 nets. 
Ethereum's token networks have an average of 442788/1192722 nodes/edges. To maintain a reasonable computation time, we obtain a subgraph via the maximum weight subgraph approximation method of \cite{MaxSubgraph:Vassilevska:2006}, which allows to reduce the dynamic network size considering only most $M$ active edges and its corresponding nodes. In these experiments, we use dynamic networks with $N=100$ nodes. Let $\mathcal{G}_t = \{\mathcal{V}_t, \mathcal{E}_t,\Tilde{W}_t\}$ denotes the reduced Ethereum blockchain network on day $t$ and $X_t \in \mathbb{R}^{N\times 1}$ be the node feature matrix, we assume a solely node feature: the node degree. Each node in $\mathcal{V}_t$ is a buyer/seller and edges in $\mathcal{E}_t$ represent transactions in the network. To construct the similarity matrix $\Tilde{W}_t$, the normalized number of transactions between node pairs $(u,v)$ serves as the edge weight value $w_{t, uv}\in\Tilde{W}_t$.

\subsection{Experiment Settings}
For multi-step time series forecasting, we evaluate the performances of Z-GCNETs on 4 time series datasets versus 13 state-of-the-art baselines (SOAs). Among them, Historical Average (HA) and Vector Auto-Regression (VAR)~\cite{hamilton2020time} are the statistical time series models. FC-LSTM~\cite{sutskever2014sequence} and GRU-ED~\cite{cho2014learning} are RNN-based neural networks. DSANet~\cite{huang2019dsanet} is the self-attention networks. DCRNN~\cite{li2017diffusion}, STGCN~\cite{stgcn}, GraphWaveNet~\cite{graphwavenet}, ASTGCN~\cite{guo2019attention}, MSTGCN~\cite{guo2019attention}, STSGCN~\cite{song2020spatial} are the spatial-temporal GCNs. AGCRN~\cite{bai2020adaptive} and LSGCN~\cite{LSGCN} are the GRU-based GCNs. We conduct our experiments on NVIDIA GeForce RTX 3090 GPU card with 24GB memory. The PeMSD4 and PeMSD8 are split in chronological order with 60\% for training sets, 20\% for validation sets, and 20\% for test sets. For PeMSD4 and PeMSD8, Z-GCNETs contains 2 layers, with each layer has 64 hidden units. We consider the window size $\tau = 12$ and horizon $h = 12$ for Z-GCNETs on both PeMSD4 and PeMSD8 datasets. Besides, the inputs of PeMSD4 and PeMSD8 are normalized by min-max normalization approach. 
\begin{table}[h!] 
\caption{The computation cost for the generation of zigzag persistence image (ZPI) and a single training epoch of Z-GCNETs.\label{running_time}}
\setlength\tabcolsep{2pt}
\centering
\begin{tabular}{lccc}
\toprule
\multirow{2}{*}{\textbf{Dataset}} & \multirow{2}{*}{\textbf{Window Size}} & \multicolumn{2}{c}{\textbf{Average Time Taken (sec)}} \\
& & ZPI & Z-GCNETs (epoch) \\
\midrule
Decentraland & 7 & 0.03  &2.09  \\
Bytom & 7 & 0.03  &2.05  \\
PeMSD4 & 12 &0.86  & 30.12 \\
PeMSD8 & 12 &0.65   &36.76  \\
\bottomrule
\end{tabular}
\end{table}
We split Bytom and Decentraland with 80\% for training sets and 20\% for test sets. For token networks, Z-GCNETs contains 2 layers, where each layer has 16 hidden units. We use one week historical data to predict the next week's data, i.e., window size $\tau = 7$ and horizon $h = 7$ over Bytom and Decentralnad datasets. All reported results are based on the weight rank clique filtration.
More detailed description of the experimental settings can be found in Appendix~A, while the analysis of sensitivity with respect to the choice of filtration is in Appendix~B. The code is available at https://github.com/Z-GCNETs/Z-GCNETs.git.

Table~\ref{running_time} reports the average running time of ZPI generation and training time per epoch of our Z-GCNETs model on all datasets.

\subsection{Comparison with the Baseline Methods}
Table~\ref{forecast_result} shows the comparison of our proposed Z-GCNETs and SOAs for traffic flow forecasting tasks. We assess model performance with Mean Absolute Error (MAE), Root Mean Square Error (RMSE), and Mean Absolute Percentage Error (MAPE) on PeMSD4 and PeMSD8. From Table~\ref{forecast_result}, we find that our proposed model Z-GCNETs consistently outperforms SOAs on PeMSD4 and PeMSD8. The improvement gain of Z-GCNETs over the next most accurate methods ranges from 0.44\% to 2.06\% in RMSE for PeMSD4 and PeMSD8. Table~\ref{token_result} shows the performance results on Bytom and Decentraland using RMSE. We find that Z-GCNETsfor the weight rank clique filtration  outperforms AGCRN by margins of 3.42\% and 2.94\% (see the results in Appendix~B for other types of filtrations). 
In contrast with SOAs, Z-GCNETs fully leverages the topological information by incorporating zigzag topological features via CNN on topological space. Given the continuous nature of time series data, our analysis and experiments show that establishing a connection between the time zigzag pairs is naturally a perfect fit to topological spaces and continuous maps.

\begin{table}[h!]
\centering
\setlength\tabcolsep{1.5pt}
\caption{Forecasting results (MAPE) on Ethereum token networks.\label{token_result}}
\begin{tabular}{lcc}
\toprule
\textbf{Model} & \textbf{Bytom} & \textbf{Decentraland} \\
\midrule
FC-LSTM~\cite{sutskever2014sequence}& 40.72\%&33.46\%\\
DCRNN~\cite{li2017diffusion}& 35.36\% &27.69\%\\
STGCN~\cite{stgcn}& 37.33\% & 28.22\%\\
GraphWaveNet~\cite{graphwavenet} &39.18\% &37.67\%\\
ASTGCN~\cite{guo2019attention}& 34.49\%&27.43\%\\
AGCRN~\cite{bai2020adaptive} &34.46\%& 26.75\% 
\\
LSGCN~\cite{LSGCN} &34.91\%&28.37\%
\\
\midrule
\textbf{Z-GCNETs} & {\bf 31.04\%} & {\bf 23.81\%}  \\
\bottomrule
\end{tabular}
\end{table}

\subsection{Ablation Study}
To better understand the importance of the different components in Z-GCNETs, we conduct ablation studies on PeMSD4 and PeMSD8 and the results are presented in Table~\ref{ablation_architecture}. The results show that Z-GCNETs have better performance over Z-GCNETs without zigzag persistence representation learning (zigzag learning), spatial graph convolution (GCN\textsubscript{Spatial}), or temporal graph convolution (GCN\textsubscript{Temporal}). Specifically, we observe that when removing GCN\textsubscript{Temporal}, the multi-step forecasting is affected significantly, i.e., Z-GCNETs outperforms Z-GCNETs without temporal graph convolution with relative gain 6.46\% on RMSE for PeMSD4. Comparison results on PeMSD8, w/o zigzag learning and w/o show the necessity for encoding topological information and modeling spatial structural information in multi-step forecasting over spatial-temporal time series dataset. Additional results for the ablation study on Ethereum tokens are presented in Appendix~C.
\begin{table}[h!]
\centering
\setlength\tabcolsep{3pt}
\caption{Ablation study of the network architecture. [*] means the GCN\textsubscript{Spatial} is only applied to the most recent time point in the sliding window.\label{ablation_architecture}}
\begin{tabular}{llccc}
\toprule
&\textbf{Architecture} & \textbf{MAE} & \textbf{RMSE} & \textbf{MAPE} \\
\midrule
\multirow{4}{*}{\textbf{PeMSD4}}&\textbf{Z-GCNETs} & {\bf 19.50} & {\bf 31.61} &{\bf 12.78\%} \\
&W/o Zigzag learning&19.65&31.94&13.01\%\\
&W/o GCN\textsubscript{Spatial}$^{*}$&19.86&31.96&13.19\%\\
&W/o GCN\textsubscript{Temporal}&20.76&33.18&13.60\%\\
\midrule\midrule
\multirow{4}{*}{\textbf{PeMSD8}}&\textbf{Z-GCNETs} & {\bf 15.76} & {\bf 25.11} &{\bf 10.01\%} \\
&W/o Zigzag learning&17.16&27.06&10.77\%\\
&W/o GCN\textsubscript{Spatial}$^{*}$&16.92&26.86&10.33\%\\
&W/o GCN\textsubscript{Temporal}&16.66&26.44&10.39\%\\
\bottomrule
\end{tabular}
\end{table}

\subsection{How does time-aware zigzag persistence help?}
To track the importance of $p$-dimensional topological features in Z-GCNETs (i.e., 0-dimensional and 1-dimensional holes), we evaluate the performance of Z-GCNETs on two different aspects: (i) the sensitivity of Z-GCNETs to different dimensional topological features and (ii) the effects of threshold $\nu_{*}$ in constructed input networks along with zigzag persistence. Table~\ref{diff_zigzag} summarizes the results using different dimensional topological features and different thresholds on PeMSD4 and PeMSD8. Under the same scale parameter $\nu_{*}$, we find that 1-dimensional topological features consistently outperform 0-dimensional terms on both datasets. Furthermore, the forecasting results on PeMSD4 are not significantly affected by varying $\nu_{*}$. However, on on PeMSD8 1-dimensional topological features constructed under $\nu_{*}$ of 0.3 yield better results than 1-dimensional terms constructed under $\nu_{*} = 0.5$.
\begin{table}
\centering
\setlength\tabcolsep{3.5pt}
\caption{Results of zigzag persistence on the dynamic network with different dimensional features and threshold values $(\nu_{*})$.\label{diff_zigzag}}
\begin{tabular}{llcccc}
\toprule
&\multirow{2}{*}{\textbf{Zigzag module}}& \multicolumn{3}{c}{\textbf{PeMSD4}}\\
\cmidrule(lr){3-5}&& MAE & RMSE & MAPE \\
\midrule
\multirow{4}{*}{\textbf{Z-GCNETs }\color{red}{+}}
&0-th ZPI$_{\nu_{*} = 0.3}$ &19.73&32.04&12.93\% \\
&1-st ZPI$_{\nu_{*} = 0.3}$&19.47  &31.66  &12.75\%\\
&0-th ZPI$_{\nu_{*} = 0.5}$&19.78 &32.20 &12.98\%\\
&1-st ZPI$_{\nu_{*} = 0.5}$&19.50&31.61&12.78\%\\
\midrule\midrule
&\multirow{2}{*}{\textbf{Zigzag module}}& \multicolumn{3}{c}{\textbf{PeMSD8}}\\
\cmidrule(lr){3-5}&& MAE & RMSE & MAPE \\
\midrule
\multirow{4}{*}{\textbf{Z-GCNETs }\color{red}{+}}
&0-th ZPI$_{\nu_{*} = 0.3}$ & 17.14 &27.24  &10.66\% \\
&1-st ZPI$_{\nu_{*} = 0.3}$&15.76 &25.11 &10.01\%\\
&0-th ZPI$_{\nu_{*} = 0.5}$ &17.22  &27.41 &10.77\% \\
&1-st ZPI$_{\nu_{*} = 0.5}$&16.77 & 26.62 &10.39\%\\
\bottomrule
\end{tabular}
\end{table}

\subsection{Robustness Study}
To assess robustness of Z-GCNETs under noisy conditions, we consider adding Gaussian noise into 30\% of training sets. The added noise follows zero-mean i.i.d Gaussian density with fixed variance ${\varsigma}^2$, i.e., 
$\mathcal{N}(0, {\varsigma}^2)$, where $\varsigma \in \{2, 4\}$. In Table~\ref{robustness_study} we report comparisons with two competitive baselines (AGCRN and LSGCN) on Decentraland and PeMSD4 using two different noise levels. Table~\ref{robustness_study} shows the performance of Z-GCNETs and two SOAs under described noisy conditions. We can see that the performance of all methods decays slowly with respect to the Gaussian noises. Despite that, we can see that Z-GCNETs is still consistently more robust than SOAs on both Decentraland and PeMSD4. On the other hand, for influence shown in Table~\ref{robustness_study}, all methods have relative lower RMSE, proving that the Gaussian noises are usually less effective for graph convolution-based models. 

\begin{table}[h!]
\centering
\setlength\tabcolsep{1pt}
\caption{Robustness study on Decentraland and PeMSD4 (RMSE).\label{robustness_study}}
\begin{tabular}{llccc}
\toprule
&\textbf{\footnotesize Noise} & \textbf{\footnotesize AGCRN} & \textbf{\footnotesize LSGCN} & \textbf{\footnotesize Z-GCNETs (ours)} \\
\midrule
\multirow{2}{*}{\textbf{\footnotesize Decentraland}}&$\mathcal{N}(0,2)$ &27.69 &36.10 &{\bf 24.12} \\
&$\mathcal{N}(0,4)$&28.12&36.79&{\bf 25.03}\\
\midrule\midrule
\multirow{2}{*}{\textbf{\footnotesize PeMSD4}}&$\mathcal{N}(0,2)$ &32.24 &34.16&{\bf 31.95}\\
&$\mathcal{N}(0,4)$&32.67&34.75&{\bf 32.18}\\
\bottomrule
\end{tabular}
\end{table}

\section{Conclusion}

Inspired the recent call for developing time-aware deep learning mechanisms by 
the US Defense Advanced Research Projects Agency (DARPA), we have proposed 
a new time-aware zigzag topological layer (Z-GCNETs) for time-conditioned GCNs. Our idea is based on the concepts of zigzag persistence whose utility remains unexplored not only in conjunction with time-aware GCN but DL in general. The new Z-GCNETs layer allows us to track the salient time-aware topological characterizations of the data persisting over time. Our results on spatio-temporal graph structured data have indicated that integration of the new time-aware zigzag topological layer into GCNs results both in enhanced forecasting performance and substantial robustness gains. 


\section{Acknowledgements}
The project has been supported in part by the grants NSF DMS 1925346, NSF ECCS 2039701, and NSF ECCS 1824716.

\bibliography{ZigzagGCNETs_ref,refV4}
\bibliographystyle{icml2021}

\appendix
\section{Additional Experimental Settings}
On PeMSD4 and PeMSD8, we train our model using Adam optimizer with an initial learning rate $lr = 0.003$ and decay rate of $\rho = 0.3$; whilst we set learning rate $lr$ to 0.001 and decay rate $\rho$ to 0.1 in Bytom and Decentraland datasets. The length of Laplacianlink is set to 2 and 3 for transportation networks and token networks, respectively. Our Z-GCNETs is trained with batch sizes of 64 and 8 on PeMSD4 and PeMSD8, respectively. On Ethereum token networks, we set the batch size to 8. We run the experiments for 300 epochs and 100 epochs on transportation networks and Ethereum token networks, respectively. In all experiments, we set the grid size of ZPI to $100 \times 100$ and use CNN model to learn zigzag persistence representation. The CNN model consists of 2 CNN layers with number of filter set to 8, kernel size to 3, stride to 2, and the global max-pooling with the pool size of $5 \times 5$. 

\section{The Choice of Filtration}
We now have also run experiments on the impact of the filtration choice. In addition to
the weight rank clique filtration, we consider
for power and weighted-degree sublevel filtrations. Table~\ref{filtration_study} shows a subset of illustrative results for Ethereum token networks. For sparser graphs such as Bytom, all filtrations tend to yield similar results. For more heterogeneous dynamic graphs with a richer topological structure, e.g., Decentraland, power filtration is the winner as it better captures evolution of the underlying graph organization.
The proposed methodology is compatible with any filtration.
\begin{table}[h!]
\caption{Z-GCNETs (MAPE) for different zigzag filtrations.\label{filtration_study}}
\setlength\tabcolsep{5pt}
\centering
\begin{tabular}{lccc}
\toprule
\textbf{Filtration}& \multicolumn{2}{c}{\textbf{Weighted-degree sublevel set}} & {\textbf{Power}} \\
\cmidrule(lr){2-3}\cmidrule(lr){4-4}
\textbf{Dataset$\backslash$Scale} & Transaction & Volume & Volume\\
\midrule
Bytom &30.56 & 30.80 & 30.79\\
Decentraland &25.18 & 24.93 & 22.15\\
\bottomrule
\end{tabular}
\end{table}

\section{Ablation study on Ethereum token networks}
To make sure that all the components of the Z-GCNETs perform well, we also conduct ablation study on Ethereum token networks. Table~\ref{ablation_study_blockchain} summarizes the results obtained on Bytom and Decentraland. The results demonstrate that our Z-GCNETs outperforms Z-GCNETs without zigzag persistence representation learning (zigzag learning), spatial graph convolution (GCN\textsubscript{Spatial}), and temporal graph convolution (GCN\textsubscript{Temporal}).

\begin{table}[h!]
\caption{Ablation study (MAPE) of Ethereum token networks. \label{ablation_study_blockchain}}
\centering
\begin{tabular}{lcc}
\toprule
\multirow{2}{*}{\textbf{Architecture}} & \multicolumn{2}{c}{\textbf{Dataset}} \\
& Bytom & Decentraland \\
\midrule
\textbf{Z-GCNETs} & {\bf 31.04\%}  &{\bf 23.81\%}  \\
W/o Zigzag learning &  33.19\%  &24.24\%  \\
W/o GCN\textsubscript{Spatial} & 34.32\%  & 25.22\% \\
W/o GCN\textsubscript{Temporal} & 31.25\%  &24.62\% \\
\bottomrule
\end{tabular}
\end{table}

\end{document}